\documentclass[10pt,twocolumn,letterpaper]{article}

\usepackage{color}
\usepackage{xspace}
\usepackage{multirow}
\usepackage{graphicx}
\usepackage{overpic}
\usepackage{cancel}

\def\ib{\boldsymbol}

\definecolor{blue}{rgb}{0,0,1}
\definecolor{red}{rgb}{1,0,0}
\definecolor{green}{rgb}{0,.5,0}
\definecolor{orange}{rgb}{0.75, 0.4, 0}
\definecolor{teal}{rgb}{0.0, 0.4, 0.4}
\definecolor{purple}{rgb}{0.65,0,0.65}

\usepackage{cvpr}              
\usepackage{graphicx}
\usepackage{caption}
\usepackage[title]{appendix}

\definecolor{cvprblue}{rgb}{0.21,0.49,0.74}
\usepackage[pagebackref,breaklinks,colorlinks,allcolors=cvprblue]{hyperref}

\def\paperID{10086} 
\def\confName{CVPR}
\def\confYear{2025}

\begin{document}

\title{ArticulatedGS: Self-supervised Digital Twin Modeling of Articulated Objects using 3D Gaussian Splatting}



\author{Junfu Guo \textsuperscript{1}\hspace{-0.2cm} \and
Yu Xin\textsuperscript{1}\hspace{-0.2cm} \and
Gaoyi Liu\textsuperscript{1}\hspace{-0.2cm} \and
Kai Xu\textsuperscript{2}\hspace{-0.2cm} \and
Ligang Liu \textsuperscript{1}\hspace{-0.2cm} \and
Ruizhen Hu \textsuperscript{3}\thanks{Corresponding author} \and
\textsuperscript{1}University of Science and Technology of China \\
\textsuperscript{2}National University of Defense Technology \\
\textsuperscript{3}Shenzhen University\\
}
\maketitle

\begin{abstract}
We tackle the challenge of concurrent reconstruction at the part level with the RGB appearance and estimation of motion parameters for building digital twins of articulated objects using the 3D Gaussian Splatting (3D-GS) method. With two distinct sets of multi-view imagery, each depicting an object in separate static articulation configurations, we reconstruct the articulated object in 3D Gaussian representations with both appearance and geometry information at the same time. Our approach decoupled multiple highly interdependent parameters through a multi-step optimization process, thereby achieving a stable optimization procedure and high-quality outcomes. We introduce ArticulatedGS, a self-supervised, comprehensive framework that autonomously learns to model shapes and appearances at the part level and synchronizes the optimization of motion parameters, all without reliance on 3D supervision, motion cues, or semantic labels. Our experimental results demonstrate that, among comparable methodologies, our approach has achieved optimal outcomes in terms of part segmentation accuracy, motion estimation accuracy, and visual quality. The code will be made publicly available at our website \url{https://guojunfu-tech.github.io/articulatedGS-io/}
\end{abstract}

%
%


\begin{figure}[t]
  \centering
      \begin{overpic}[width=\linewidth]{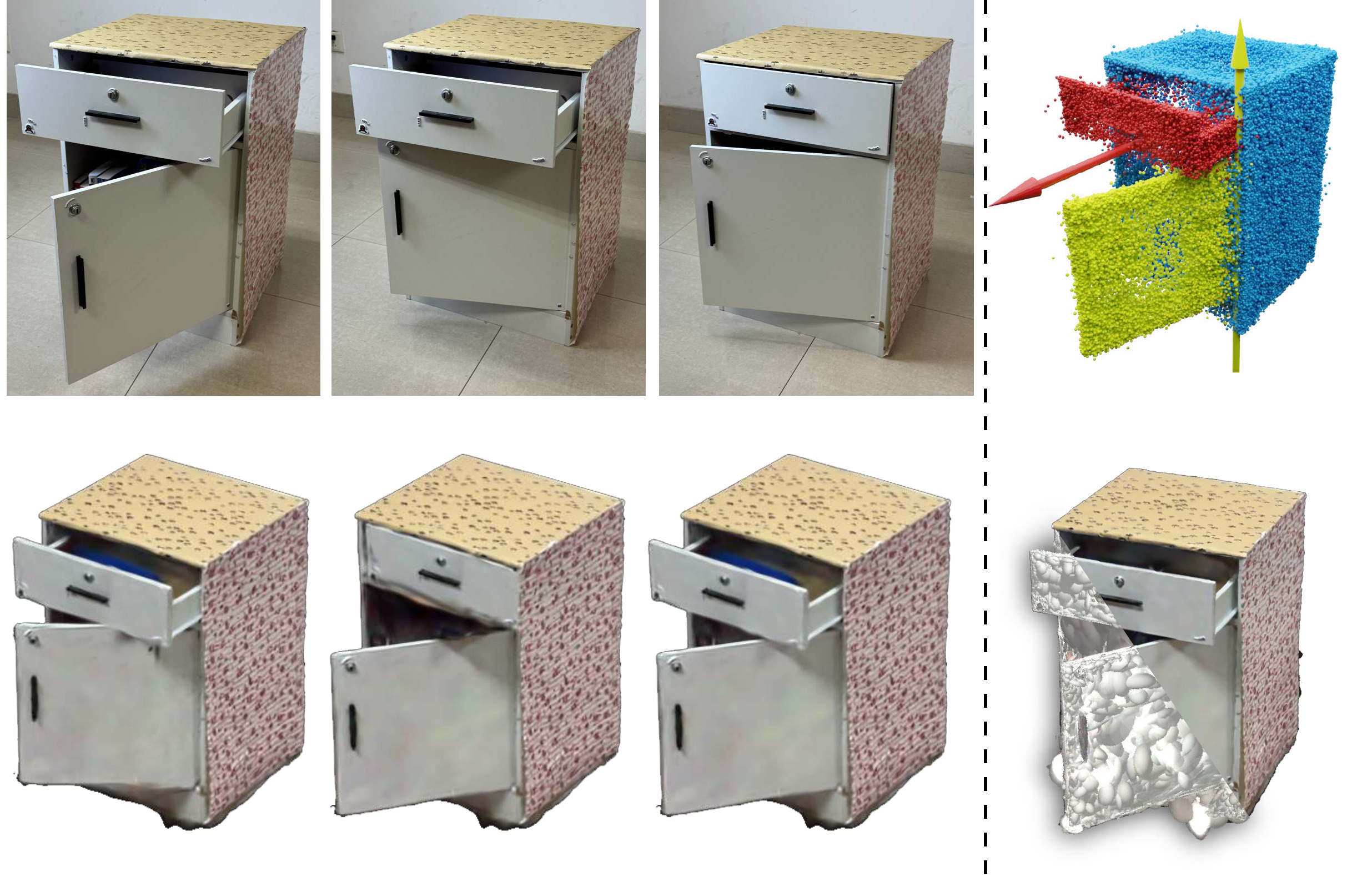}
        \put(22,32){\scriptsize \textbf{Real-World Object}}
        \put(22,1){\scriptsize \textbf{Any-State Synthesis}}
        \put(75,34){\scriptsize \textbf{Motion Estimation}}
        \put(73,31){\scriptsize \textbf{ \& Part Segmentation}}
        \put(79,1){\scriptsize \textbf{3D Gaussians}}
    \end{overpic}
    \caption{\textbf{ArticulatedGS} is a self-supervised modeling and reasoning pipeline to reconstruct the articulated objects using 3D Gaussian Splatting.}
     \label{fig:teaser}
\end{figure}

\section{Introduction}

Articulated objects, such as laptops, ovens, and drawers, are a part of our daily life. 
Building the digital twin of an articulated object with accurate reconstruction of and fine-grained analysis often plays an important role in robotics~\cite{qian2023understanding,mo2021where2act, zhang2024manidext, lou2024robo}, 
animation~\cite{yang2022banmo,li2023animatable}, and 
simulation~\cite{yang2022object,xiang2020sapien}. 
Moreover, it is also desirable to equip the digital twins with color appearance to facilitate downstream applications such as training sim-to-real transferable embodied AI~\cite{szot2021habitat,deitke2022️, shridhar2020alfred}. Consequently, the capturing and reconstruction of \textit{shape geometry}, \textit{visual appearance}, and \textit{articulation parameters} is indispensable in building accurate and functional digital twins of articulated objects. 

The problem of modeling high-fidelity digital twins of articulated objects is receiving increasing attention in recent years~\cite{liu2024survey,liu2023paris,jiang2022ditto,sharf2014mobility}.
Most of the previous works~\cite{jiang2022ditto, liu2023paris, lei2023nap} are concentrated on the regeneration or construction of 3D geometrical representation of articulated objects, which may involve the estimation of articulation parameters or functionalities to animate the objects into different articulation states. 
The result is usually a mesh-based representation of articulated objects with relatively low geometric fidelity and visual realism.
Moreover, the reliance on supervised learning, which typically consumes a large 3D dataset with articulation annotation, also limits the generalizability to unseen object categories.
To our knowledge, the only \textit{self-supervised} work that takes RGB observations as input is PARIS \cite{liu2023paris}. PARIS adopts the neural radiance field (NeRF) as an intermediate representation and outputs a mesh with compromised accuracy of geometry and appearance. Additionally, it exhibits instability and suboptimality issues in the optimization process.
%

Recently, there has also been a massive success of 3D Gaussian Splatting (3D-GS) \cite{kerbl3Dgaussians} in 3D scene learning and novel view synthesis as well as extensions to various applications such as 3D reconstruction~\cite{dhamo2023headgas, lei2023gart,qian2023gaussianavatars} and dynamic scene rendering~\cite{duan20244d,gao2024gaussianflow}. 
3D-GS requires only RGB images as input and enables rapid scene reconstruction with real-time and realistic rendering. The reconstructed scene is composed of numerous Gaussians  densely distributed around the object's surface, thereby allowing for modeling rich geometric details.
Recent works~\cite{yan2024street,xie2023physgaussian, lu2024manigaussian} make it possible to interact with digital twins represented with 3D-GS for physically meaningful simulation and manipulation. The fast and realistic rendering of 3D-GS also benefits learning in simulation. 
For those purposes, what has been missing is to reconstruct digital twins of articulated objects with the representation of 3D Gaussians.
In this work, we aim to take the RGB observations of two states of a given articulated object as an input and reconstruct a 3D-GS model of the object with \textit{geometry}, \textit{appearance} and \textit{mobility}.

Optimizing for object geometry (masks of movable parts), appearance, and motion parameters simultaneously from only two-view inputs is a highly ill-posed problem. This is because the three terms are deeply intertwined. The object's movable parts, undergoing rigid body transformations based on motion parameters, can greatly impact on the learning of part geometry and visual appearance. The same goes for the opposite way; the optimization of the movable mask and the motion parameters can be significantly affected by rendering quality.
To that end, we propose to decouple the optimization process into multiple sub-processes. We first reconstruct the 3D-GS model of the start state. We then train a network to estimate a deformation flow for moving the movable Gaussians and initialize the geometry structure (movable mask) for the end state. Based on the initial movable mask, we leverage the geometry structure to supervise the learning of the motion parameters. At last, we jointly update the appearance, motion parameters, and movable mask to obtain the final reconstruction of the articulated object.

Evaluation on an extended dataset from the previous works~\cite{geng2023gapartnet,liu2023paris} demonstrates that our method achieves the state-of-the-art performance of visual quality, motion parameter estimation, and part segmentation. Our method also shows enhanced efficiency in model training and more stability of the optimization process compared to the previous self-supervision methods.
Our work makes the following contributions:
\begin{itemize}
    \item A 3D Gaussian model for building digital twins of articulated objects in a self-supervised fashion;
    \item A progressive optimization scheme to the challenging optimization task with efficiency and stability;
    \item New state-of-the-art results of reconstruction quality, motion estimation, and visual quality for articulated objects.
\end{itemize}

\begin{figure*}[htbp]
    \centering
    \includegraphics[width=\linewidth]{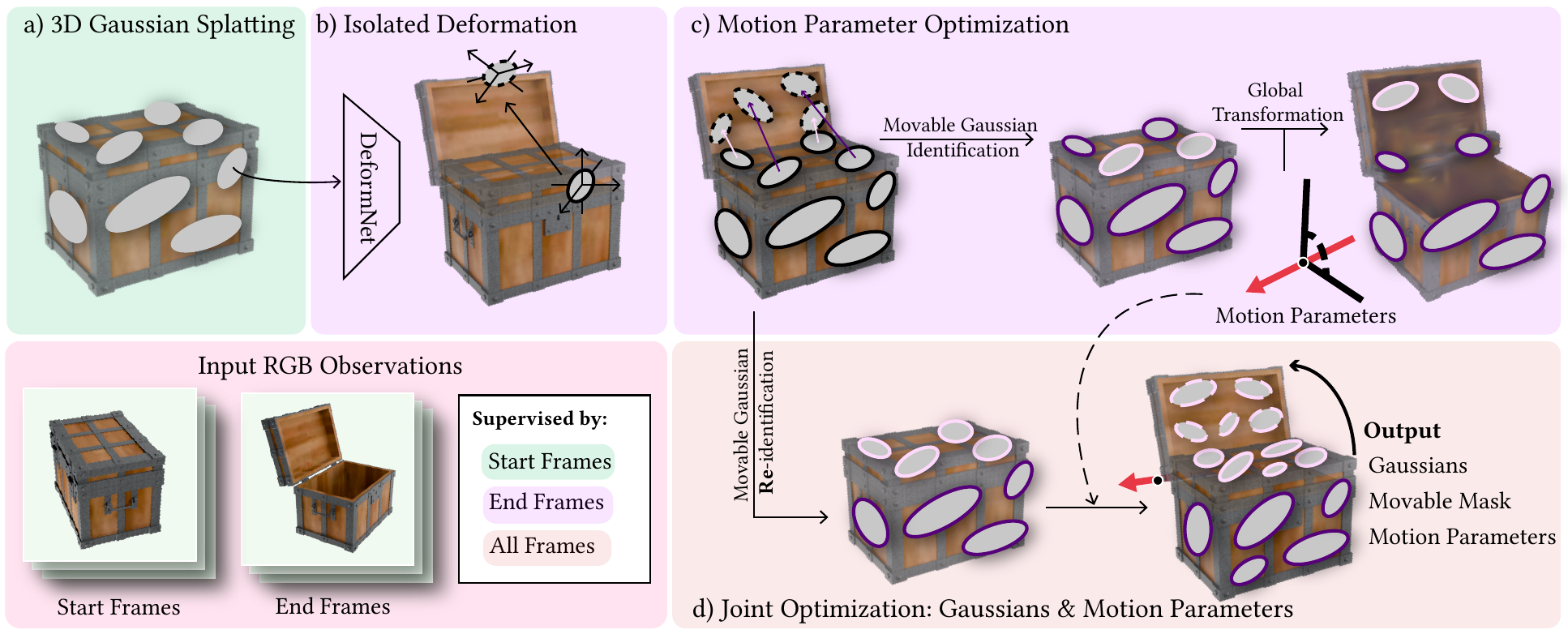}
    \caption{A pipeline overview of our self-supervised 3D Articulated Objects modeling method. (a) Given the sets of RGB observation sequence of two states $I_0$ and $I_1$ of an arbitrary articulated object, we first reconstruct the start state using 3D Gaussian Splatting method into the initial Gaussians $\mathcal{G}=\{g_i\}_{i=1}^N$ . 
   (b) Each Gaussian $g_i$ is then fed into DeformNet, where its Gaussian kernel's 3D coordinates $\ib{x}_i$ is input to predict its individual rigid motion $\{\delta\ib{x}_i, \delta\ib{r}_i\}$ within a local coordinate system, with the end state images $I_1$ serving as visual supervision. 
  (c) Based on the transformation predicted by DeformNet, we classify the displacements after regularization to obtain an initial moving Gaussian, and then apply the motion parameters to this moving Gaussian to perform a global rigid transformation, which is supervised by $I_1$, ultimately resulting in a motion category and converged motion parameters. 
   (d) Subsequently, we optimize the preprocessed Gaussians, motion parameters, and the movable mask in tandem to achieve the final results.
   }
\label{pipeline}
\end{figure*}

\section{Related work}

\paragraph{Articulated Modeling.}
The primary objective of articulated object reconstruction is to accurately recover the three-dimensional (3D) geometry of articulated entities.
Pekelny and Gotsman \cite{pekelny2008articulated} introduced one of the pioneering works in this area, which focuses on reconstructing the surface from a sequence of point clouds. 
A-SDF \cite{mu2021sdf} represents an innovative approach, the inaugural study to reconstruct articulated objects from signed distance fields (SDFs) derived from a solitary snapshot. This technique permits the deformation of objects into any conceivable articulation state. While echoing the goals of A-SDF, the work of \cite{wei2022self} diverges in their approach by utilizing multi-view RGB images for reconstruction purposes. CARTO \cite{heppert2023carto} and \cite{le2024articulate} advances this concept further by incorporating a single stereo RGB image as input. 
A common limitation across these three studies is their tendency to reconstruct the articulated object in its entirety as a unified surface. 
To counteract this limitation, CLANeRF \cite{tseng2022cla}, Ditto \cite{jiang2022ditto}, PARIS \cite{liu2023paris}, LEIA \cite{swaminathan2024leia}, and Deng's work \cite{deng2024articulate} have each developed methodologies that reconstruct articulated objects at the part level. CLANeRF, PARIS and Deng utilize multi-view RGB images as their input, whereas Ditto processes point cloud data. 
To overcome the instability of the optimization process of PARIS, the DTA \cite{weng2024neural} and ArtGS\cite{liu2025building} take depth as extra information to achieve the same goal but higher performance. 
This work utilizes multiple modules and incorporates prior knowledge such as image feature extraction, but the output of this method does not inlcude appearance.
In contrast, our approach requires only RGB data and employs a fully self-supervised method, with an optimization time that is approximately one-third of theirs, yet still achieves similar quality results.
In our work, we adopt the same RGB series input as PARIS and solve the stabilization problem while yielding a more accurate result with higher visualization quality. Moreover, the Gaussian representation we utilize can, compared to PARIS, quite naturally encompass both appearance and geometry characteristics.
For more complete coverage of works in this field, please see the recent survey \cite{liu2024survey}.

\paragraph{Deformable 3D Gaussians.}
Gaussian Splatting \cite{kerbl3Dgaussians} offers improved rendering quality and speed for radiance fields. Several concurrent works have extended 3D-GS from representing static scenes to deformation areas. 
\cite{luiten2023dynamic,wu20234d,duisterhof2023md,kratimenos2023dynmf,lin2023gaussian,yang2023deformable,jiang20233d} use the deformation field to represent the GS deformation with time varies. Given the RGB observations at dense time frames, the deformation field takes the position of each GS and the time frame and predicts the deformation of each GS in position, rotation, and scaling. 
The work of \cite{gao2024mesh} introduced a mesh-based approach to combine the mesh with 3D Gaussians, leveraging the mesh to guide the splitting process and enhancing the overall quality of the learned GS.
However, these methods are highly dependent on the density of frames; when frames are sparse, the Gaussian motion process between frames cannot be accurately predicted.
In our work, the movable Gaussians deform together with the motion parameters, remaining visually and geometrically consistent between the two states.


Another line of works focuses on using 3D-GS to model the organic articulated objects, such as human \cite{jiang20233d,lei2023gart, kocabas2024hugs}, body parts \cite{jiang20233d,dhamo2023headgas,pokhariya2024manus}, and animals \cite{lei2023gart}. The motion allowed for these objects is usually structurally complex, with a relatively large number of joints varied in the topology of connections and degrees of freedom. However, the kinematic structure is relatively fixed and need to be predefined for a specific species. 

\section{Method}

\subsection{Problem Statement}
In this work, we take two sets of multi-view RGB images $\mathcal{I}_0 = \{I_0^i\}$ and $\mathcal{I}_1=\{I_1^i\}$, together with the corresponding camera poses, as input to represent the start and end states of the target articulated object, respectively. Our goal is to use a set of 3D Gaussians associated with part segmentation and articulated motion parameters to reconstruct such an articulated object. 

To formally define the problem, we first explain how the part-aware 3D Gaussians and the motion parameters are represented, and then define the optimization objective.


\paragraph{Part-aware 3D Gaussian Representation.}
In the default setting, 
we assume that there is only one \textit{movable} part during the movement between two states, and the rest of the object is \textit{unmovable}, and use a movable mask to distinguish those two types of Gaussians. 
In more details, the classical 3D Gaussian can be represented as $\mathcal{G}=\{g_i=(\boldsymbol{x}_i, \boldsymbol{r}_i, \boldsymbol{s}_i, \sigma_i, sh_i)|i=1...N\}$, where each Gaussian $g$ has a center position $\boldsymbol{x}$, the rotation matrix represented by a quaternion $\ib{r}\in\mathbb{SO}(3)$, $S$ as a scaling matrix represented by a 3D vector $\ib{s}$, and view-dependent appearance represented by spherical harmonics (SH) parameters $sh$. In our method, for each $g_i$, we add an extra element $m_i$ to present the movable mask such that the $g_i$ is a movable part if $m_i\neq 0$. 
Therefore, our part-ware 3D Gaussian representation $\mathcal{G}$ can be denoted as $\mathcal{G}=\{\mathcal{G}_m, \mathcal{G}_u\}$, where $\mathcal{G}_m = \{g_i|m_i\neq0\}$ and $\mathcal{G}_u$ represents the complement. 

\paragraph{Articulated Motion Representation.}
Following previous works\cite{li2020category,jiang2022ditto,liu2023paris}, we consider two types of joints: \textit{prismatic} and \textit{revolute}. The prismatic joint is parameterized by a 3D translation axis $\boldsymbol{a}\in\mathbb{R}^3, ||\ib{a}||=1$ and a move distance $d$ along this axis; And the revolute joint is parameterized by a 3D rotation orientation axis $\boldsymbol{l}\in\mathbb{R}^3, ||\ib{l}||=1$, a pivot $\boldsymbol{p}\in\mathbb{R}^3$ that locates the axis in the world coordinates, and the rotate angle $\theta$.

With the motion parameters above, we can construct rigid-body transformation functions for revolute joint on each Gaussian by $(\hat{\ib{x}},\hat{\ib{r}})=f_{\ib{l},\ib{p},\theta}(\ib{x}, \ib{r})$, and prismatic joint, $\hat{\ib{x}} = f_{\ib{a},d}(\ib{x})$, allowing each Gaussian $g(\ib{x}, \ib{r})$ to calculate its post-motion $\hat{g}(\hat{\ib{x}}, \hat{\ib{r}})$. In this way, the transformed set of Gaussians can be represented as 
\begin{equation}
    \hat{\mathcal{G}} = \{\mathcal{G}_u, f(\mathcal{G}_m)\}, 
\end{equation}
where $f\in\{f_{\ib{l},\ib{p},\theta}, f_{\ib{a},d}\}$. 


\paragraph{Optimization formulation. }
With the part-aware Gaussians and motion parameters, our goal is to optimize them all together to achieve high-fidelity rendering of images similar to the input RGB images across different angles and states. 
Thus, our optimization problem can be formulated as  
\begin{equation}
    \underset{\mathcal{G}, f}{\operatorname{argmin}}\sum_i{\mathcal{L}_{app}\big(I_0^i(\mathcal{G})-I_0^i\big)+\sum_j\mathcal{L}_{app}\big(I_1^j(\hat{\mathcal{G}})-I_1^j\big)},
\label{eq:obj}
\end{equation}
where $I(\mathcal{G})$ represents the image rasterized by Gaussian $\mathcal{G}$ with the camera extrinsic and intrinsic of image $I$, and $\mathcal{L}_{app}$ is the same appearance loss function used in the 3D Gaussian Splatting method~\cite{kerbl3Dgaussians}: 
\begin{equation}
    \mathcal{L}_{app} = (1-\lambda)\mathcal{L}_1 + \lambda\mathcal{L}_{D\_SSIM}.
\label{eq:app}
\end{equation}

\subsection{Overview}

Given the complex problem formulated in Eq. (\ref{eq:obj}), optimizing the motion parameters, the movable mask, and the Gaussian representations concurrently can result in a tendency to settle for a local optimum or to face computational failure. To address this, we introduce a novel process that methodically tackles the overarching objective in a step-wise fashion, each step aimed at securing initial parameter estimates that are proximal to the ground truth, thus enhancing the likelihood of reaching the optimal solution.

The pipeline of our method is illustrated in Fig. \ref{pipeline}. 
We first train a set of static Gaussians $\mathcal{G}$ with only viewpoints $I_0$ from the start frames.
Then, we use a deformation network to predict each Gaussian's translation and rotation, supervised by $I_1$. With the deformed Gaussians, we classify the movements and obtain the initial movable Gaussians $\mathcal{G}_m$ and unmovable Gaussians $\mathcal{G}_u$. 
After that, We optimize the motion parameters in $\mathbb{SE}(3)$ by casting the rigid movement to $\mathcal{G}_m$. During this step, we classify the motion type according to the motion parameters. 
With the pre-processed motion parameters and movable mask, we jointly optimize the Gaussians' representation, the movable mask, and the motion parameters to obtain the final result. 
Below we will explain the details of each step.

\begin{figure}[tbp]
    \centering
    \includegraphics[width=\linewidth]{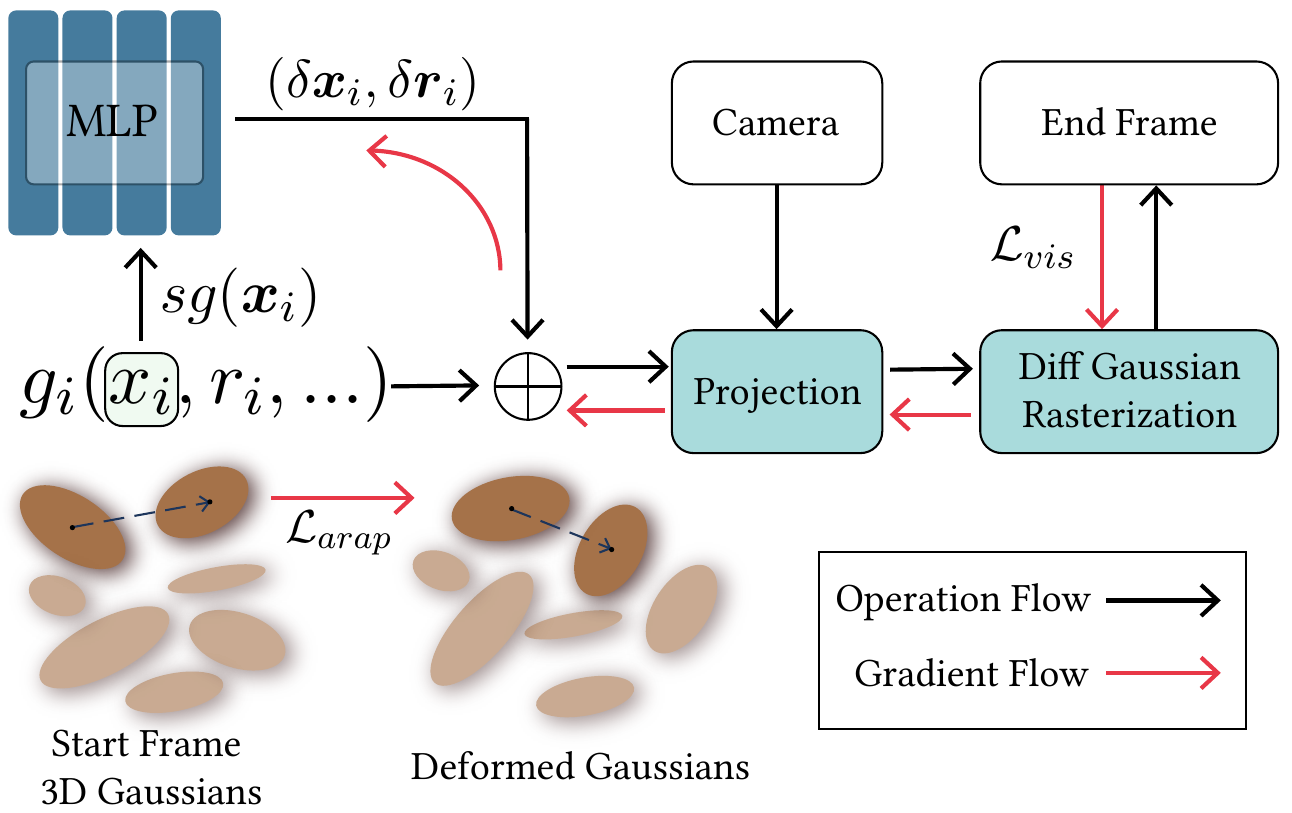}
    \caption{ The process diagram of DeformNet illustrates our approach where a neural network is utilized to predict the rigid transformation of a Gaussian. The input consists of the kernel's positions $\ib{x}_i$ for each Gaussian $g_i$, and the output is the displacement and relative rotation to the end state. In addition to visual supervision by $I_1$, we also impose geometrical supervision $\mathcal{L}_{arap}$ that maintains local consistency among the local Gaussians.}
    \label{deform}
\end{figure}

\subsection{Isolated Deformation Prediction}
Fig.~\ref{deform} gives an illustration of our isolated deformation Process. 
With the trained Gaussians of start frames, we do not optimize the motion parameters directly, but use a 
network
$f_{\ib{\theta}}$ to predict the deformation field of the Gaussians. 
The deformation field takes the position of each 3D Gaussian $\ib{x}_i$ as input, and output $\delta \ib{x}_i$ and $\delta \ib{r}_i$, which represent the translation and rotation of Gaussian $\ib{g}_i$:
\begin{equation}
    (\delta \ib{x}_i, \delta \ib{r}_i) = f_{\ib{\theta}}(sg(\ib{x}_i)).
\end{equation}
Note that as we do not want to change the origin location of each Gaussian but only predict the deformation field, we add a stop-gradient operation $sg(\cdot)$ to $\ib{x}_i$. 
The network architecture is relatively simple, which consists of a four-layer one-dimensional Convolutional Network (1D CNN) connected to a four-layer Multi-Layer Perceptron (MLP). 

The deformed 3D Gaussians $\hat{\mathcal{G}}=\{\hat{g}_i=(\hat{\ib{x}}_i, \hat{\ib{r}_i},...)|\hat{\ib{x}}_i=\ib{x}_i+\delta \ib{x}_i, \hat{\ib{r}}_i=\ib{r}_i+\delta \ib{r}_i\}$ is then put into the differential Gaussian rasterization pipeline to get the rendered image $\hat{I}=I(\hat{\mathcal{G}})$. The appearance loss $\mathcal{L}_{app}$ (defined in Eq.~(\ref{eq:app})) is then computed between the rendered image $\hat{I}$ and the target image $I_1$.

Moreover, with the prior that the desired deformation should be part-wise rigid, we add an extra as-rigid-as-possible (ARAP) loss $\mathcal{L}_{arap}$ as in previous works \cite{luiten2023dynamic, huang2023sc} to encourage the motion to be locally rigid:
\begin{gather}
    \mathcal{L}_{arap} = \frac{1}{k|\mathcal{S}|}\sum_{i\in\mathcal{S}}\sum_{j\in knn_{i,k}} \mathcal{L}_{arap}^{i,j}, \\
    \mathcal{L}_{arap}^{i,j} = \omega_{i,j}||(\ib{x}_j - \ib{x}_i) - R_i \hat{R}_i^{-1}(\hat{\ib{x}}_j - \hat{\ib{x}}_i)||_2.
\end{gather}
Here, we restrict the set of Gaussians $j$ to be the k-nearest-neighbor of $i$ $(k=20)$, and the weight factor for the Gaussian pair is defined as:
\begin{equation}
    \omega_{i,j} = exp\big(-\lambda_\omega ||\ib{x}_j-\ib{x}_i||_2^2\big)
\end{equation}
where $\lambda_\omega=20$ in our experiments. 

Therefore, the final loss function is the combination of the appearance loss and the ARAP loss:
\begin{equation}
    \mathcal{L}_{deform} = \mathcal{L}_{app} + \lambda_{arap}\mathcal{L}_{arap},
\end{equation}
where $\lambda_{arap}=1$ in our experiments.



\subsection{Motion Parameter Optimization}

Once we obtain the deformation field of all the Gaussians, we can first roughly identify the movable part and estimate the motion parameters based on their deformations. Note that the movable part doesn't need to be very accurate at this step, as the estimated motion parameters could be further refined in the following joint optimization, and our goal here is to get an good approximation as initialization for joint optimization.

To identify the movable part, we first normalize the displacement $||\delta \ib{x}||_2$ within 0 and 1, and take the Gaussians with $\delta \ib{x} \geq \tau$ as the movable Gaussians $\mathcal{G}_m$.
Based on our observations, selecting a lower threshold $\tau$ often results in the inclusion of stationary Gaussians, which can lead to sub-optimal outcomes when optimizing motion parameters later on. 
Thus, we tend to set a higher threshold to get a partial set of movable parts with higher confidence, which is usually already enough for motion parameter prediction for the entire part. 
We set $\tau=0.3$ in our experiments to classify the movable Gaussians.

With the movable Gaussians $\mathcal{G}_m$,
we proceed to optimize the motion parameters by applying a global rigid transformation to $\mathcal{G}_m$.
We observed that when fitting a prismatic motion with revolute motion parameters, the pivot of the revolute parameters typically moves a considerable distance to ensure that the movable Gaussians undergo minimal rotation during the motion. 
Therefore, we initiate with optimizing the revolute parameters. 
The motion is considered to be prismatic if the pivot $||\ib{p}||>R$, where $R=1$ in our experiments, and we will reinitialize the optimization parameters and the motion function to align with the prismatic motion.

We utilize both appearance and geometric supervision to refine the motion parameters: 
\begin{equation}
\mathcal{L}_{param}=\mathcal{L}_{app}+\mathcal{L}_{geo}.
\end{equation}
For the appearance, we maintain the use of the end frame for supervision, and thus $\mathcal{L}_{app}$ is the same as the one used for the Deformation Network. 
For the geometric supervision, we preserve the Gaussian points that have undergone transformation by the Deformation Network and calculate the Chamfer distance between the coordinates of these two sets of points to provide geometric guidance. Therefore, the geometric loss is defined as:
\begin{equation}
    \mathcal{L}_{geo}=d_{c}\big(f(\ib{x}), \ib{x}+\delta \ib{x}\big), f\in\{f_{\ib{l},\ib{p},\theta}, f_{\ib{a},d}\},
\end{equation}
where $d_{c}(\cdot, \cdot)$ denotes the chamfer distance.

\subsection{Final Joint Optimization}
With the good initialization of 3D Gaussians, movable mask, and the motion parameters, now we are able to perform a joint optimization to find optimal solution.

\begin{table*}
\centering
\resizebox{\linewidth}{!}{
\begin{tabular}{ccc|cccccccccc|cc} 
\toprule
\multicolumn{3}{c|}{\multirow{2}{*}{}}                             & \multicolumn{10}{c|}{Simulation~ ~}                                                                                                                                   & \multicolumn{2}{c}{Real}   \\
\multicolumn{3}{c|}{}                                              & Foldchair      & Fridge         & Laptop*        & Oven*          & Scissor        & Stapler        & USB            & Washer         & Blade         & Storage*       & Fridge         & Storage*         \\ 
\hline
\multirow{12}{*}{Motion}   & \multirow{4}{*}{Ang Err}  & Ditto    & 89.35          & 89.30          & 3.12           & 0.96           & 4.50           & 89.86          & 89.77          & 89.51          & 79.54         & 6.32          & 1.71           & \textbf{5.88}            \\
                           &                            & PARIS    & 7.90           & 9.19           & \textbf{0.02}           & \textbf{0.04}  & 3.92           & 0.73          & \textbf{0.13}  & 25.18          & 15.18         & \textbf{0.03} & \textbf{1.64}  & 43.13           \\
                           &                            & Vanilla  & 87.47          & 14.08          & 11.65          & 44.95          & 89.80          & 84.85          & 75.50          & 84.69          & 88.76         & 83.17         & 82.99          & 78.35           \\
                           &                            & Ours~    & \textbf{0.02}  & \textbf{0.04}  & 0.07  & \textbf{0.04}  & \textbf{0.20}  & \textbf{0.03}  & 0.30           & \textbf{0.06}  & \textbf{1.99} & 0.05          & 5.41           & 41.52  \\ 
\cline{2-15}
                           & \multirow{4}{*}{Pos Err}  & Ditto    & 3.770          & 1.020          & 0.010          & 0.130          & 5.700          & 0.200          & 5.410          & 0.660          & -             & -             & 1.840          & -               \\
                           &                            & PARIS    & 0.374          & 0.298          & 0.010          & \textbf{0.003} & 2.154          & 2.258          & 2.367          & 1.502          & -             & -             & 0.340          & -               \\
                           &                            & Vanilla  & 0.036          & 0.503          & 0.265          & 0.055          & 0.027          & 0.127          & 0.013          & 0.030          & -             & -             & 0.084          & -               \\
                           &                            & Ours~    & \textbf{0.003} & \textbf{0.001} & \textbf{0.003} & 0.007          & \textbf{0.002} & \textbf{0.000} & \textbf{0.000} & \textbf{0.004} & -             & -             & \textbf{0.057} & -               \\ 
\cline{2-15}
                           & \multirow{4}{*}{Geo Dist} & Ditto    & 99.36          & F              & 5.18           & 2.09           & 19.28          & 56.61          & 80.60          & 55.72          & F             & \textbf{0.09}          & 8.43           & \textbf{0.38}            \\
                           &                            & PARIS    & 131.82         & 24.64          & \textbf{0.03}  & \textbf{0.04}  & 120.61         & 110.71         & 64.91          & 60.62          & 0.54          & 0.14 & \textbf{2.16}  & 0.56            \\
                           &                            & Vanilla  & 80.00          & 59.49          & 37.92          & 29.97          & 40.00          & 40.05          & 44.77          & 30.00          & 48.41         & 32.55         & 35.45          & 48.25           \\
                           &                            & Ours~    & \textbf{0.05}  & \textbf{0.18}  & 0.55           & 0.61           & \textbf{0.15}  & \textbf{0.06}  & \textbf{0.33}  & \textbf{0.34}  & \textbf{0.09} & 0.30          & 9.38           & 0.41   \\ 
\hline
\multirow{12}{*}{Geometry} & \multirow{4}{*}{CD-s}      & Ditto    & 33.79          & 3.05           & \textbf{0.25}  & 2.52           & 39.07          & 41.64          & 2.64           & 10.32          & 46.90         & 9.18          & 47.01          & \textbf{16.09}  \\
                           &                            & PARIS    & 9.12           & 3.73           & 0.45           & 12.85          & 1.83           & 1.96           & 2.58           & 25.19          & 1.33          & 12.80         & 42.57          & 54.54           \\
                           &                            & Vanilla  & 308.39         & 8.81           & 389.67        & F              & F              & 100.23         & 88.12          & 312.80         & F             & 218.01        & F              & F               \\
                           &                            & Ours~    & \textbf{0.33}  & \textbf{0.90}  & 2.96           & \textbf{2.06}  & \textbf{0.34}  & \textbf{1.68}  & \textbf{1.02}  & \textbf{6.17}  & \textbf{0.69} & \textbf{2.96} & \textbf{37.01} & 50.12           \\ 
\cline{2-15}
                           & \multirow{4}{*}{CD-d}      & Ditto    & 141.11         & 0.99           & \textbf{0.19}           & 0.94           & 20.68          & 31.21          & 15.88          & 12.89          & 195.93        & 2.20          & 50.60          & \textbf{20.35}  \\
                           &                            & PARIS    & 8.79           & 7.76           & 0.49           & 28.51          & 46.69          & 19.36          & 5.53           & 178.39         & 25.29         & 76.75         & 45.66          & 864.82          \\
                           &                            & Vanilla~ & 189.26         & 254.50         & 159.95         & 717.63         & 90.31          & 116.08         & 88.12          & 512.02         & 102.91        & 854.10        & 923.22         & 965.15          \\
                           &                            & Ours~    & \textbf{0.32}  & \textbf{0.59}  & 6.23  & \textbf{0.80}  & \textbf{0.41}  & \textbf{1.14}  & \textbf{0.90}  & \textbf{0.17}  & \textbf{4.75} & \textbf{0.71} & \textbf{43.00} & 730.45          \\ 
\cline{2-15}
                           & \multirow{4}{*}{CD-w}      & Ditto    & 6.80           & 2.16           & \textbf{0.31}  & 2.51           & 1.70           & 2.38           & 2.09           & 7.29           & 42.04         & 3.91          & \textbf{6.50}           & \textbf{14.08}  \\
                           &                            & PARIS    & 1.90           & 2.53           & 0.50           & 1.94          & 10.20          & 3.60           & 2.31           & 24.71          & 0.44          & 8.34          & 22.98          & 63.35           \\
                           &                            & Vanilla  & 5.91           & 9.99           & 4.73           & 2.80           & 7.63           & 33.34          & 8.74           & 8.37           &        5.52       & 3.21          & 30.44          & 78.65           \\
                           &                            & Ours~    & \textbf{0.36}  & \textbf{0.85}  & 0.87           & \textbf{1.91}  & \textbf{0.33}  & \textbf{1.42}  & \textbf{0.93}  & \textbf{5.64}  & \textbf{0.34} & \textbf{2.09} & 13.25          & 55.57           \\
\bottomrule
\end{tabular}}
\caption{
The results on the PARIS dataset, encompassing both synthetic and real data, are presented. Objects marked with a `*' are the categories that Ditto \cite{jiang2022ditto}  has been trained to recognize. Occasionally, Ditto may provide incorrect motion type predictions, which are indicated with an `F' for joint state and a `*' for joint axis or position. There is a chance that Vanilla may fail to successfully segment the object, classifying the entire object as part of the movable category, which we also mark with an `F'. 
}
\label{table:results}
\end{table*}

Given that we already obtained quite accurate motion parameters in the previous step, we employ a lower threshold $\tau=0.1$ to reclassify the Gaussians to get a more strict movable mask for the joint optimization. 
With this movable mask, we divide the Gaussians into movable part $\mathcal{G}_m$  and unmovable part $\mathcal{G}_u$. 
We apply the motion parameters obtained earlier to the movable part to reach its corresponding end state, while for the unmovable part, the start and end states are the same.
Note that during the densification and pruning processes of the Gaussians, inaccuracies in the classification within two masks can be treated as noise within the Gaussians. This noise will be automatically updated alongside the Gaussians' densification and pruning processes, thereby achieving an accurate movable mask.
We employ multi-perspective images from both states to supervise the rendering results before and after the movable part's movement, with the expectation that the loss values for each state will eventually converge to the same magnitude. 



\begin{figure}[tbp]
    \centering
    \begin{overpic}[width=\linewidth]{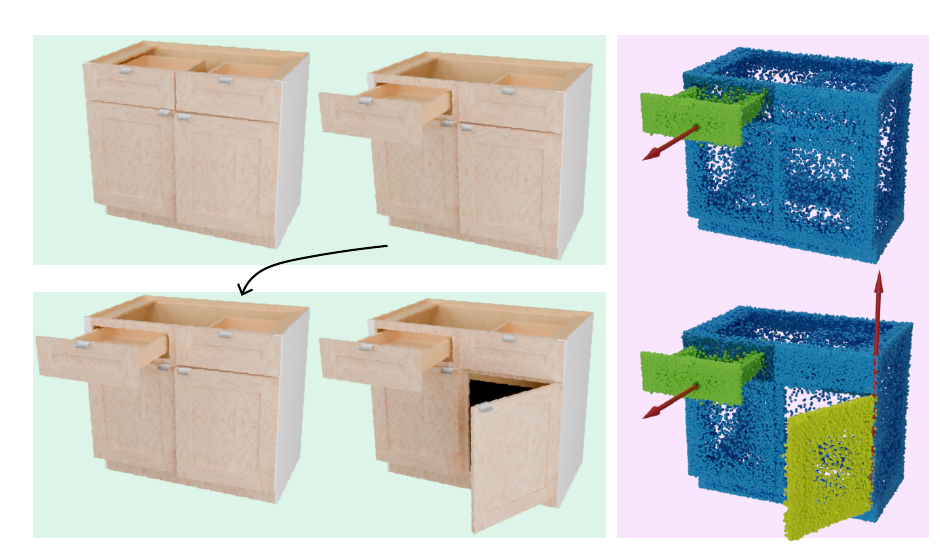}
         \put(13,55){Start State}
         \put(45,55){End State}
         \put(77,55){End State}
         \put(0,40){\rotatebox{90}{Step 1}}
         \put(0,10){\rotatebox{90}{Step 2}}
         \put(30,27){Same}
    \end{overpic}
        \caption{An example of modeling a multi-part articulated object. Taking the target object in multiple states, we update the Gaussians sequentially to reconstruct the final object with multiple sets of movable parts.}
        \label{fig:multi}
\end{figure}

\subsection{Multi-Part Object Modeling}
To extend our method to reconstruct objects with multiple moving parts, we dissect the entire procedure into multiple sub-processes of single-object modeling and address each in turn, as shown in Fig. \ref{fig:multi}.

Throughout the reconstruction process, we consistently maintain the same set of Gaussians. 
We align the start state of the subsequent sub-process with one of the states from the preceding sub-process, ensuring that only an additional moving component is introduced in the new sub-process. 
Before entering the next sub-process, we configure the movable components within the existing Gaussian to the starting state of the subsequent sub-process using the motion parameters derived from the previous one. Within each sub-process, we employ a new identifier to label the movable mask while concurrently optimizing the unmovable Gaussians to update the new movable mask. 
Given that our movable mask is capable of being simultaneously densified and pruned with the Gaussian, the Gaussians marked by the movable mask from the previous step can also be optimized in the new sub-process. Furthermore, there is an advantage in that for the new sub-process, we can bypass the step of pre-training the initial Gaussian and proceed directly to the deformation network. 

\section{Results and Evaluation}

\subsection{Experimental Setup}

\paragraph{Baselines.}
We compare our method to most related SOTA works Ditto \cite{jiang2022ditto} and PARIS \cite{liu2023paris}. Note that we follow PARIS's protocol and report the results derived from Ditto’s released pre-trained model, which was trained on four object categories from the Shape2Motion dataset \cite{wang2019shape2motion}.
Moreover, we have also constructed a trivial reconstruction method based on our framework, denoted as \textit{Vanilla}. Rather than employing our multi-step optimization process, we simultaneously optimize all parameters. 
We directly initialize two sets of Gaussians, one representing the static part and the other the mobile part. To simplify the problem, we also assume that the type of motion is known. 

\paragraph{Evaluation metrics.}
To evaluate the efficacy of our methodology, we have designed a comparative experimental framework that spans three critical aspects: reasoning of motion, precision of geometry, and visual quality of novel views, as in previous works~\cite{liu2023paris, weng2024neural}. The details of all the metrics are provided in the supplementary material.
Moreover, we compare our approach's computational efficiency and memory utilization against PARIS, which is also a self-supervised method like ours.

\begin{figure}[tbp]
    \centering
    \begin{overpic}[width=\linewidth]{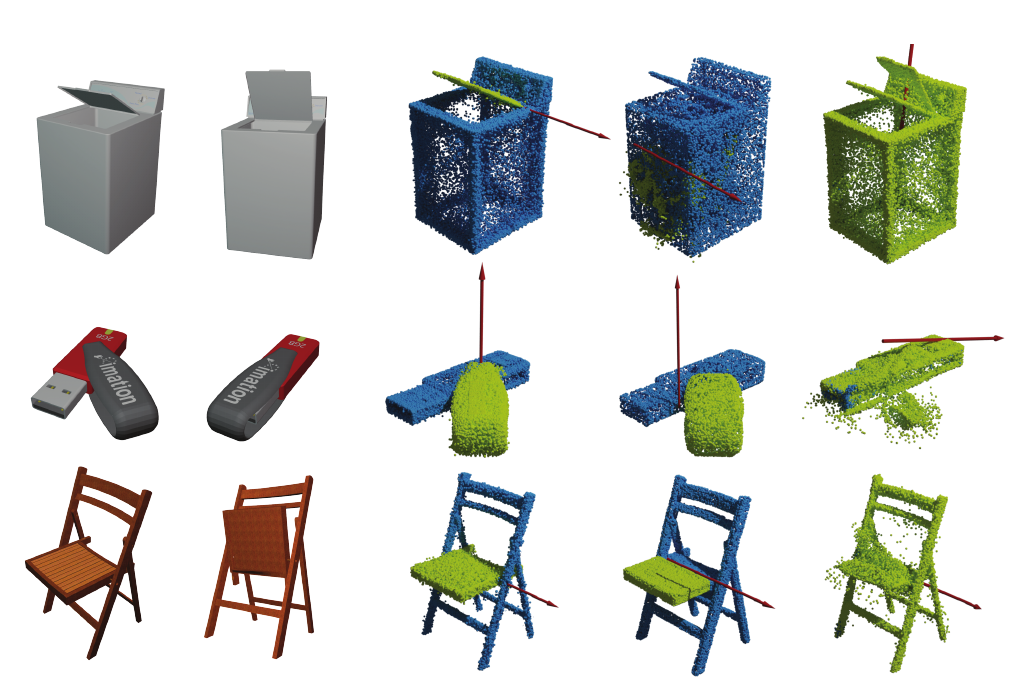}
    \put(2,62){\small \textbf{Start State}}
    \put(20,62){\small \textbf{End State}}
    \put(44,62){\small \textbf{Ours}}
    \put(63,62){\small \textbf{PARIS}}
    \put(84,62){\small \textbf{Vanilla}}
    \end{overpic}
    \caption{Qualitative results of part segmentation and motion estimation for some of the synthetic objects derived from the PARIS dataset. }
    \label{fig:motion_and_seg}
\end{figure}

\begin{table*}
\centering

\resizebox{0.8\linewidth}{!}{
\begin{tabular}{c|ccccccccccc} 
\toprule
Metrics                  & Methods      & Foldchair      & Fridge         & Laptop         & Oven           & Scissor        & Stapler        & USB            & Washer         & Blade          & Storage         \\ 
\hline
\multirow{3}{*}{PSNR-s ($\uparrow$)}  & PARIS        & 37.95          & 36.45          & 36.59          & 35.59          & 37.64          & 33.84          & 37.58          & 37.22          & 38.28          & 35.02           \\
                         & Vanilla & 31.49          & 23.21          & 22.45          & 24.16          & 21.08          & 22.26          & 21.99          & 26.65          & 23.87          & 29.65           \\
                         & Ours~        & \textbf{41.68} & \textbf{36.89} & \textbf{37.08} & \textbf{36.60} & \textbf{40.99} & \textbf{43.36} & \textbf{41.09} & \textbf{40.63} & \textbf{41.31} & \textbf{38.65}  \\ 
\hline
\multirow{3}{*}{PSNR-e ($\uparrow$)}  & PARIS        & 38.45          & 33.68          & 34.24          & 34.33          & 38.15          & 40.19          & 36.13          & 37.75          & 37.13          & 36.41           \\
                         & Vanilla & 31.42          & 25.44          & 22.56          & 23.38          & 21.07          & 22.75          & 23.11          & 26.09          & 27.76          & 29.65           \\
                         & Ours~        & \textbf{42.83} & \textbf{37.63} & \textbf{35.37} & \textbf{36.72} & \textbf{41.13} & \textbf{42.84} & \textbf{42.27} & \textbf{39.98} & \textbf{41.36} & \textbf{38.08}  \\ 
\hline
\multirow{3}{*}{SSIM-s ($\uparrow$)}  & PARIS        & 0.956          & 0.968          & 0.987          & 0.966          & 0.953          & 0.957          & 0.979          & 0.970          & 0.988          & 0.961           \\
                         & Vanilla     & 0.949          & 0.938          & 0.936          & 0.954          & 0.961         & 0.964          & 0.926          & 0.979          & 0.940          & 0.952           \\
                         & Ours~        & \textbf{0.980} & \textbf{0.987} & \textbf{0.988} & \textbf{0.984} & \textbf{0.995} & \textbf{0.996} & \textbf{0.993} & \textbf{0.993} & \textbf{0.996} & \textbf{0.976}  \\ 
\hline
\multirow{3}{*}{SSIM-e ($\uparrow$)}  & PARIS        & 0.955          & 0.965          & 0.984          & 0.970          & 0.954          & 0.956          & 0.980          & 0.972          & 0.987          & 0.965           \\
                         & Vanilla & 0.943          & 0.948          & 0.924          & 0.945          & 0.933          & 0.967          & 0.931          & 0.980          & 0.947          & 0.950           \\
                         & Ours~        & \textbf{0.983} & \textbf{0.989} & \textbf{0.986} & \textbf{0.984} & \textbf{0.995} & \textbf{0.995} & \textbf{0.994} & \textbf{0.992} & \textbf{0.996} & \textbf{0.975}  \\ 
\hline
\multirow{3}{*}{LPIPS-s ($\downarrow$)} & PARIS        & 0.062          & 0.038          & 0.047          & 0.070          & 0.067          & 0.064          & 0.070          & 0.020          & 0.008          & 0.084           \\
                         & Vanilla & 0.084          & 0.083          & 0.083          & 0.081          & 0.083          & 0.081          & 0.087          & 0.084          & 0.081          & 0.114           \\
                         & Ours~        & \textbf{0.056} & \textbf{0.039} & \textbf{0.036} & \textbf{0.040} & \textbf{0.012} & \textbf{0.008} & \textbf{0.019} & \textbf{0.021} & \textbf{0.008} & \textbf{0.081}  \\ 
\hline
\multirow{3}{*}{LPIPS-e ($\downarrow$)} & PARIS        & 0.061          & 0.039          & 0.047          & 0.044          & 0.016          & 0.021          & 0.027          & 0.020          & 0.009          & 0.083           \\
                         & Vanilla & 0.094          & 0.073          & 0.090          & 0.073          & 0.067          & 0.051          & 0.068          & 0.037          & 0.085          & 0.124           \\
                         & Ours~        & \textbf{0.059} & \textbf{0.033} & \textbf{0.031} & \textbf{0.039} & \textbf{0.012} & \textbf{0.009} & \textbf{0.017} & \textbf{0.023} & \textbf{0.007} & \textbf{0.087}  \\
\bottomrule
\end{tabular}}
 \caption{Comparison of visual quality on the PARIS dataset, where (-s) and (-e) denote the \textit{start} and \textit{end} states. 
}
\label{table:vis_t}
\end{table*}
\begin{figure}[htbp]
    \centering
    \begin{overpic}[width=\linewidth]{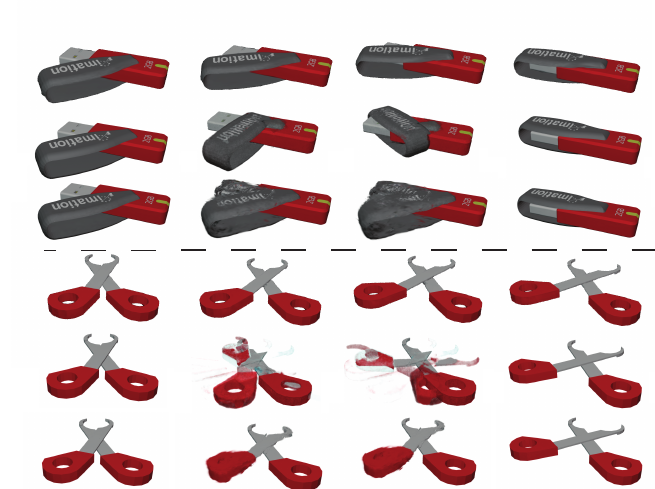}
    \put(6,70){\small \textbf{Start State}}
    \put(80,70){\small \textbf{End State}}
    
    
    \put(1,38){\small \rotatebox{90}{DGS}}
    \put(1,46){\small \rotatebox{90}{PARIS}}
    \put(1,60){\small \rotatebox{90}{Ours}}
    
    \put(1,27){\small \rotatebox{90}{Ours}}
    \put(1,11){\small \rotatebox{90}{PARIS}}
    \put(1,0){\small \rotatebox{90}{DGS}}
    \end{overpic}
    \caption{Illustration of unseen states inference. 
    }
    \label{fig:seq}
\end{figure}
\begin{figure*}[tbp]
    \centering
    \label{multi}
    \begin{overpic}[width=\linewidth]{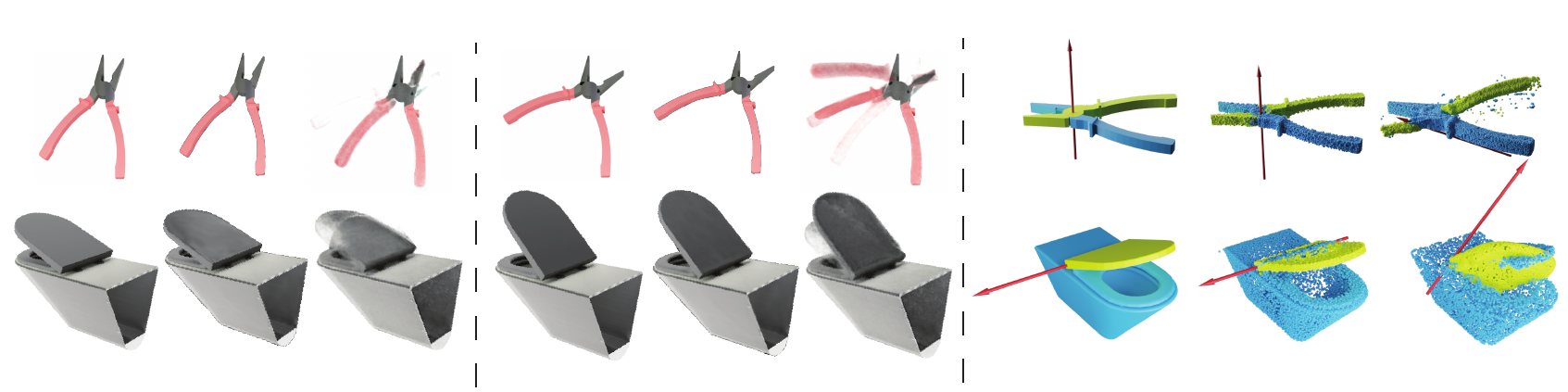}
    \put(11,22){\small \textbf{Start State}}
    \put(43,22){\small \textbf{End State}}
    \put(79,22){\small \textbf{Results}}
    
    \put(4,0){\small GT}
    \put(13,0){\small Ours}
    \put(22,0){\small PARIS}

    \put(36,0){\small GT}
    \put(44,0){\small Ours}
    \put(54,0){\small PARIS}

    \put(67,0){\small GT}
    \put(80,0){\small Ours}
    \put(92,0){\small PARIS}

    \end{overpic}
    \caption{Illustration of the effects of applying our method and the PARIS algorithm to our dataset. 
}
    \label{fig:our_data}
\end{figure*}
\begin{figure*}[tbp]
    \centering

    \begin{overpic}[width=0.9\linewidth]{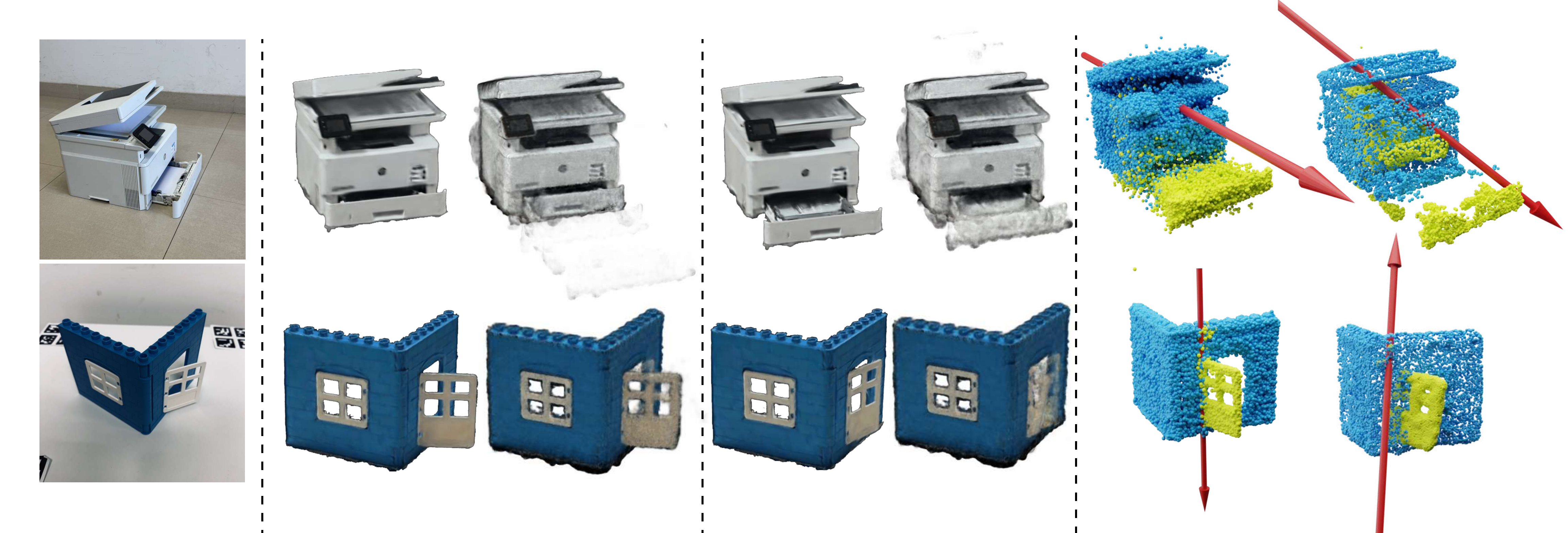}
        \put(1,32){\small \textbf{Real-World Object}}
        \put(26,32){\small \textbf{Start State}}
        \put(51,32){\small \textbf{End State}}
        \put(84,32){\small \textbf{Results}}
        
        \put(22,1){\small Ours}
        \put(34,1){\small PARIS}
        \put(49,1){\small Ours}
        \put(60,1){\small PARIS}
        \put(74,1){\small Ours}
        \put(89,1){\small PARIS}
    \end{overpic}
    \caption{Illustrations of our method applied to the reconstruction results of real-world objects.}
    \label{fig:real-ours}
\end{figure*}

\subsection{Comparison to Baselines}
Table \ref{table:results} and Table \ref{table:vis_t} show results on the PARIS object dataset, including synthetic and real instances, summarized over 10 trials. Ditto's performance is contingent upon its exposure to object forms and structural templates acquired from the training dataset, which includes categories such as laptops, ovens, and storage units. The algorithm demonstrates proficiency in reconstructing shapes for objects that fall within its training purview. PARIS, on the other hand, experiences significant fluctuations in its performance across various trials. Although it occasionally achieves commendable shape reconstruction, it also suffers from occasional catastrophic failures. These inconsistencies result in a markedly inferior overall performance in both the reconstruction of shape and the delineation of object articulation. 
Our approach demonstrates resilience to varying initial conditions and consistently delivers precise reconstructions of both shape and articulation throughout multiple iterations.

Figure \ref{fig:motion_and_seg} shows some representative results of various methods in part-segmentation and motion estimation. We have visualized each of our Gaussian movable masks in the form of point clouds, which allows for a more intuitive observation of the accuracy of our method in segmentation and estimation. 
Figure~\ref{fig:seq} illustrates the predicted articulation sequence based on the reconstructions obtained by different methods. 
PARIS struggles to accurately estimate motion parameters for certain objects. Even when the segmentation results are close to correct, the motion process still deviates from the actual outcome. 
We also added the Deformable GS (DGS) \cite{yang2023deformable}, a recent and pioneering 4D Gaussian work, for visual comparison. Due to the sparsity of the input temporal states, DGS can only reconstruct the appearance of the two input states with relative accuracy. It lacks prior knowledge of articulated motion and thus cannot infer the actual transformation process between the two states. 
We also created our own datasets for comparative analysis. As shown in Figure \ref{fig:our_data}, our method demonstrates superior results across a greater variety of objects compared to previous works, and more results can be found in the supplementary material. 

Note that our method on real objects has not achieved consistently the best performance for all metrics in Table~\ref{table:results}. That is because, in the PARIS dataset, the real-world object masks have various imperfections at the edges, such as missing parts of the object or mistakenly including portions of the background as part of the object. 
Additionally, obtaining the ground truth for motion and segmentation of real-world objects is also quite peculiar, as manual annotation of motion and segmentation can easily introduce errors in the ground truth. Consequently, using it is not entirely effective for quantitative evaluation. 
To demonstrate the effectiveness of our method in reconstructing real objects, we scanned multiple objects and applied our method for digital-twin modeling, as shown in Figure \ref{fig:teaser} and Figure \ref{fig:real-ours}. 

For additional visual results, please refer to the supplementary materials.

\subsection{Ablation Studies}

The ``Vanilla" baseline in Table~\ref{table:results}, Table \ref{table:vis_t} and  
Figure~\ref{fig:motion_and_seg} refers to the framework of our method without the progressive optimization process. The results demonstrate that the visual similarity between the initial and final states is preserved even after the removal of our progressive optimization. Nonetheless, part segmentation tends to optimize directly to a scenario where all Gaussians are consolidated into a single category. Consequently, the motion parameters that are learned result in a motion that is nearly static. 


\section{Conclusion} 
In this work, we introduce ArticulatedGS, a self-supervised framework designed to build digital twins of articulated objects using 3D Gaussian Splatting.
Utilizing two distinct sets of multi-view images, each depicting an object in separate static articulation configurations, we reconstruct the articulated object in 3D Gaussian representations, integrating both appearance and mobility information simultaneously.
Our method adeptly disentangles a multitude of highly interrelated parameters through a sophisticated multi-step optimization process. This approach not only ensures a stable and reliable optimization procedure but also delivers high-fidelity results. These outcomes are crucial for the precise depiction of articulated objects within digital twins, ensuring that they are accurately and realistically represented.
Our approach also has certain limitations, such as the requirement for world coordinate alignment between two states, and corresponding areas under two different states should have similar lighting effects. More detailed analysis can be found in the supplementary material.


\section*{Acknowledgement}
 This work was supported in part by NSFC (62322207, 62025207, 62325211, 62132021), Shenzhen Science and Technology Program (RCYX20210609103121030), and the Major Program of Xiangjiang Laboratory (23XJ01009). We would like to thank Youcheng Cai and Changhao Li from University of Science and Technology of China, for their advice on the methodology. We also thank the anonymous reviewers for their valuable comments and suggestions.

{
    \small
    \bibliographystyle{ieeenat_fullname}
    \bibliography{sample-base}
}

\end{document}